\def\year{2021}\relax
\documentclass[letterpaper]{article} 

\usepackage{aaai21}  
\usepackage{times}  
\usepackage{helvet} 
\usepackage{courier}  
\usepackage[hyphens]{url}  
\usepackage{graphicx} 
\usepackage{natbib}  
\usepackage{caption} 
\usepackage{bm}
\usepackage{amssymb}
\usepackage{amsmath,amsfonts}
\usepackage{mathtools}
\usepackage{booktabs}
\usepackage{xcolor}
\usepackage[linesnumbered,ruled]{algorithm2e}
\usepackage{caption}
\usepackage{subcaption}
\usepackage{enumitem}
\setlist{leftmargin=8mm}
\usepackage{caption}
\usepackage[switch]{lineno}

\newcommand{\chu}[1]{{\color{red} #1}}
\newcommand{\nop}[1]{}

\newtheorem{assumption}{Assumption}

\newtheorem{thm}{Theorem}
\newtheorem{defi}{Definition}

\def\R{\mathbb{R}}

\def\mG{\mathcal{G}}
\def\mD{\mathcal{D}}
\def\mF{\mathcal{F}}
\def\mA{\mathcal{A}}

\title{Personalized Cross-Silo Federated Learning on Non-IID Data}

\author{Yutao Huang\textsuperscript{\rm 1}, Lingyang Chu\thanks{Lingyang Chu and Yutao Huang contribute equally in this work. The API of this work is available at \protect\url{https://developer.huaweicloud.com/develop/aigallery/notebook/detail?id=6d4a9521-6a4d-4b6d-b84d-943d7c7b1cbd}, free registration at Huawei Cloud is required before use.}\textsuperscript{\rm 2}, Zirui Zhou\textsuperscript{\rm 3}, Lanjun Wang\textsuperscript{\rm 3}, \\
Jiangchuan Liu\textsuperscript{\rm 1}, Jian Pei\textsuperscript{\rm 1}, Yong Zhang\textsuperscript{\rm 3}
}
\affiliations{

    \textsuperscript{\rm 1}Simon Fraser University, Burnaby, Canada $\mid$ yutaoh@sfu.ca, jcliu@sfu.ca, and jpei@cs.sfu.ca\\
    \textsuperscript{\rm 2}McMaster University, Hamilton, Canada $\mid$ chul9@mcmaster.ca\\
    \textsuperscript{\rm 3}Huawei Technologies Canada, Burnaby, Canada $\mid$ \{zirui.zhou, lanjun.wang, yong.zhang3\}@huawei.com\\
}

\begin{document}

\maketitle

\nop{For the challenging computational environment of IOT/edge computing, personalized federated learning allows every client to train a strong personalized cloud model by effectively collaborating with the other clients in a privacy-preserving manner.}

\begin{abstract}

Non-IID data present a tough challenge for federated learning. In this paper, we explore a novel idea of facilitating pairwise collaborations between clients with similar data. We propose FedAMP, a new method employing federated attentive message passing to facilitate similar clients to collaborate more. We establish the convergence of FedAMP for both convex and non-convex models, and propose a heuristic method to further improve the performance of FedAMP when clients adopt deep neural networks as personalized models. Our extensive experiments on benchmark data sets demonstrate the superior performance of the proposed methods.

\nop{
Personalized cross-silo federated learning has raised a lot of interest from companies and institutes who want to participate as clients to collaboratively train strong personalized machine learning models without exposing their sensitive private data.
The performance of personalized cross-silo federated learning is largely determined by the effectiveness of inter-client collaboration.
However, when the data is non-IID across all clients, it is challenging to model the pair-wise collaboration relationships between clients without knowing their private data.
In this paper, we propose to tackle this problem by a novel framework named federated attentive message passing (FedAMP). 
The key idea is to realize an attentive collaboration mechanism that adaptively discovers the underlying pair-wise collaboration relationships between clients by iteratively encouraging similar clients to collaborate more with each other.
We establish the convergence of FedAMP for both convex and non-convex models, and propose a heuristic method to further improve the performance of FedAMP when clients adopt deep neural networks as personalized models. Extensive experiments demonstrate the superior performance of the proposed methods.
}
\end{abstract}


\section{Introduction}

Federated learning~\cite{yang2019federated} facilitates collaborations among a set of clients and preserves their privacy so that the clients can achieve better machine learning performance than individually working alone. The underlying idea is to collectively learn from data from all clients. The initial idea of federated learning starts from aggregating models from clients to achieve a global model so that the global model can be more general and capable. The effectiveness of this global collaboration theme that is not differentiating among all clients highly depends on the data distribution among clients.  It works well on IID data, that is, clients are similar to each other in their private data distribution.

\nop{
With the advances in internet of things (IoT) and 5G technology, 
}
\nop{
Huge amount of edge devices (i.e., clients), such as smart phones, wearable devices and autonomous vehicles, are generating big private data in a surprisingly fast speed~\cite{poushter2016smartphone}.
Due to the ever-growing concerns and restrictions on data privacy,
many federated learning frameworks have been proposed to collaboratively conduct machine learning tasks for all clients while keeping their data privacy intact~\cite{mcmahan2016communication, li2018federated, zhao2018federated, erlingsson2014rappor, hard2018federated, yang2018applied}.
}

In many application scenarios where collaborations among clients are needed to train machine learning models, data are unfortunately not IID.  For example, consider the cases of personalized cross-silo federated learning~\cite{kairouz2019advances}, where there are tens or hundreds of clients and the private data of clients may be different in size, class distributions and even the distribution of each class.  Global collaboration without considering individual private data often cannot achieve good performance for individual clients. 

\nop{Some federated learning methods, such as FedAvg and FedProx, try to fix the problem by conducting an additional fine-tuning step after a global model is trained~\cite{ben2010theory, cortes2014domain, mansour2020three, mansour2009domain,  schneider2019mass, wang2019federated}. 
While those methods work in some cases, they cannot solve the problem systematically as demonstrated in our experimental results (e.g., dataset CIFAR100 in Table~\ref{tab:result}).}

Some federated learning methods try to fix the problem by conducting an additional fine-tuning step after a global model is trained~\cite{ben2010theory, cortes2014domain, mansour2020three, mansour2009domain,  schneider2019mass, wang2019federated}.While those methods work in some cases, they cannot solve the problem systematically as demonstrated in our experimental results (e.g., data set CIFAR100 in Table~\ref{tab:result}).

We argue that the fundamental bottleneck in personalized cross-silo federated learning with non-IID data is the misassumption of one global model can fit all clients. Consider the scenario where each client tries to train a model on customers' sentiments on food in a country.  Different clients collect data in different countries. Obviously, customers' reviews on food are likely to be related to their cultures, life-styles, and environments. Unlikely there exists a global model universally fitting all countries.  Instead, pairwise collaborations among countries that share similarity in culture, life-styles, environments and other factors may be the key to accomplish good performance in personalized cross-silo federated learning with non-IID data.

Carrying the above insight, in this paper, we tackle the challenging personalized cross-silo federated learning problem by a novel \textit{attentive message passing mechanism} that adaptively facilitates the underlying pairwise collaborations between clients by iteratively encouraging similar clients to collaborate more.
We make several technical contributions. 

We propose a novel method \emph{federated attentive message passing} (FedAMP) whose central idea is the attentive message passing mechanism.
FedAMP allows each client to own a local personalized model, but does not use a single global model on the cloud server to conduct collaborations.
Instead, it maintains a personalized cloud model on the cloud server for each client, and realizes the attentive message passing mechanism by attentively passing the personalized model of each client as a message to the personalized cloud models with similar model parameters.  Moreover, FedAMP updates the personalized cloud model of each client by a weighted convex combination of all the messages it receives. 
This adaptively facilitates the underlying pairwise collaborations between clients and significantly improves the effectiveness of collaboration.

We prove the convergence of FedAMP for both convex and non-convex personalized models. Furthermore, we propose a heuristic method to further improve the performance of FedAMP on clients using deep neural networks as personalized models. We conduct extensive experiments to demonstrate the superior performance of the proposed methods.

\section{Related Works}
\label{sec:relatedworks}

Personalized federated learning for clients with non-IID data has attracted much attention~\cite{deng2020adaptive, fallah2020personalized,  kulkarni2020survey, mansour2020three}. Particularly, our work is related to global federated learning, local customization and multi-task federated learning.

\nop{
However, it has been empirically shown to diverge on non-IID data across clients~\cite{mcmahan2016communication, li2018federated, zhao2018federated}. 

and Zhao \textit{et~al.}~\cite{zhao2018federated} globally share a small subset of data across all clients, which, however, infringes the data privacy of clients.
}

Global federated learning~\cite{ji2019learning,  mcmahan2016communication, wang2020federated,   yurochkin2019bayesian} trains a single global model to minimize an empirical risk function over the union of the data across all clients.
When the data is non-IID across different clients, however, it is difficult to converge to a good global model that achieves a good personalized performance on every client~\cite{kairouz2019advances, li2018federated, mcmahan2016communication, zhao2018federated}.

Local customization methods~\cite{ chen2018federated, fallah2020personalized, jiang2019improving, khodak2019adaptive, kulkarni2020survey, mansour2020three, nichol2018first, schneider2019mass, wang2019federated} build a personalized model for each client by customizing a well-trained global model.  There are several ways to conduct customization. 
A practical way to customize a personalized model is local fine-tuning~\cite{ben2010theory, cortes2014domain, mansour2020three, mansour2009domain,  schneider2019mass, wang2019federated}, where the global model is fine-tuned using the private data of each client to produce a personalized model for the client.
Similarly, meta-learning methods~\cite{chen2018federated, fallah2020personalized, jiang2019improving, khodak2019adaptive,  kulkarni2020survey, nichol2018first} can be extended to customize personalized models by adapting a well-trained global model on the local data of a client~\cite{kairouz2019advances}.
Model mixture methods~\cite{deng2020adaptive, hanzely2020federated} customize for each client by combining the global model with the client's latent local model.
SCAFFOLD~\cite{karimireddy2019scaffold} customizes the gradient updates of personalized models to correct client-drifts between personalized models and a global model.

Most existing local customization methods use a single global model to conduct a global collaboration involving all clients. The global collaboration framework only allows contributions from all clients to a global model and customization of the global model for each client.  It does not allow pairwise collaboration among clients with similar data, and thus may meet dramatic difficulty on non-IID data.

Smith \textit{et~al.}~\cite{smith2017federated} model the pair-wise collaboration relationships between clients by extending distributed multi-task learning to federated learning. They tackle the problem by a primal-dual optimization method that achieves great performance on convex models. At the same time, due to its rigid requirement of strong duality, their method is not applicable when clients adopt deep neural networks as personalized models.

\nop{Distributed multi-task learning~\cite{vanhaesebrouck2017decentralized, bellet2017personalized, zantedeschi2019fully, xie2017privacy} tried to conduct multi-task learning when the data for each task is separately owned by different clients.
While these methods have the potential to protect the data privacy of clients, they cannot robustly handle the unreliable operating environment of federated learning~\cite{smith2017federated}.
}

\nop{
Vanhaesebrouck \textit{et~al.}~\cite{vanhaesebrouck2017decentralized} and Bellet \textit{et~al.}~\cite{bellet2017personalized} trained personalized models by propagating and averaging models between clients using a pre-exisitng peer-to-peer collaboration network.
However, the collaboration network is usually not known apriori in practice. Zantedeschi~\cite{zantedeschi2019fully} proposed a boosting method to jointly learn the collaboration network with the personalized models of clients, but it needs to pre-train base classifiers on a third-party training data, which is hard to construct without knowing the data distributions of clients.

Xie \textit{et~al.}~\cite{xie2017privacy} proposed a distributed proximal gradient algorithm to train personalized models with guaranteed differential data privacy, but it could not robustly handle client drops.
}

Different from all existing work, our study explores pairwise collaboration among clients.  Our method is particularly effective when clients' data are non-IID, and can take the great advantage of similarity among clients.

\section{Personalized Federated Learning Problem}
In this section, we introduce the personalized federated learning problem that aims to collaboratively train personalized models for a set of clients using the non-IID private data of all clients in a privacy-preserving manner~\cite{kairouz2019advances, zhao2018federated}.

\nop{
In this section, we introduce the personalized federated learning problem
that involves a set of clients and a central cloud server.
Each client owns a local model and a set of private training data, where the training data of different clients are sampled from distinct distributions. 

Each client owns a personalized local model and a set of private training data collected from distinct distributions. 
The cloud is a central server that coordinates the training process.

effective collaborations between clients in a .

}

\nop{
, each owning a personalized local model and a set of private training data collected from distinct distributions, attempt to train their personalized models by effectively collaborating with each other in a privacy-preserving manner~\cite{kairouz2019advances, zhao2018federated}. 
}

\nop{ under the coordination of a service provider}

\nop{
, where the personalized local models of all clients are in the same type but have distinct model parameters.
}

\nop{
We will illustrate how to derive the \textbf{personalized cloud model} for each client later in this subsection.
}

\nop{
Inspired by several widely adopted formulations in multi-task learning~ \cite{han2015learning,zhang2017survey}, 
}

\nop{We say $w_i$ is the model of the client $\mathcal{C}_i$ when the context is clear.}

Consider $m$ clients $C_1, \ldots, C_m$ that have the same type of models $\mathcal{M}$ personalized by $m$ different sets of model parameters $\mathbf{w_1}, \ldots, \mathbf{w_m}$, respectively. 
Denote by $\mathcal{M}(\mathbf{w_i})$ and $D_i$ $(1 \leq i \leq m)$ the personalized model and the private training data set of client $C_i$, respectively.
These data sets are non-IID, that is, $D_1, \ldots, D_m$ are uniformly sampled from $m$ distinct distributions $P_1, \ldots, P_m$, respectively.
For each client $C_i$, denote by $\mathcal{V}_i$ the performance of $\mathcal{M}(\mathbf{w_i})$ on the distribution $P_i$.  Denote by $\mathcal{V}^*_i$ the best performance model $\mathcal{M}$ can achieve on $P_i$ by considering all possible parameter sets.

The \emph{personalized federated learning problem} aims to collaboratively use the private training data sets $D_1, \ldots, D_m$ to train the \textit{personalized models} $\mathcal{M}(\mathbf{w_1}), \ldots, \mathcal{M}(\mathbf{w_m})$ such that $\mathcal{V}_1, \ldots, \mathcal{V}_m$ are close to $\mathcal{V}^*_1, \ldots, \mathcal{V}^*_m$, respectively, and no private training data of any clients are exposed to any other clients or any third parties.

To be concrete, denote by $F_i:\R^d\rightarrow\R$ the training objective function that maps the model parameter set $\mathbf{w_i}\in\mathbb{R}^d$ to a real valued training loss with respect to the private training data $D_i$ of client $C_i$.
We formulate the personalized federated learning problem as 
\begin{equation}
\label{eq:our-form}
\min_{W} \ \left\{\mG(W)\coloneqq \sum_{i=1}^m F_i(\mathbf{w_i}) + \lambda\sum_{i<j}^m A(\|\mathbf{w_i} - \mathbf{w_j}\|^2)\right\}, 
\end{equation}
where $W = [\mathbf{w_1},\dots,\mathbf{w_m}]$ is a $d$-by-$m$ dimensional matrix that collects $\mathbf{w_1},\dots,\mathbf{w_m}$ as its columns and $\lambda>0$ is a regularization parameter.

The first term $\sum_{i=1}^m F_i(\mathbf{w_i})$ in Eq.~\eqref{eq:our-form} is the sum of the training losses of the personalized models of all clients. This term allows each client to separately train its own personalized model using its own private training data.
The second term improves the collaboration effectiveness between clients by an attention-inducing function $A(\|\mathbf{w_i} - \mathbf{w_j}\|^2)$ defined as follows.

\nop{
Here, the formulation in \eqref{eq:our-form} is a general form that represents a class of objective functions for different user-specified functions $D$.
The second term of \eqref{eq:our-form} will enforce the attentive collaboration mechanism if $D$ is properly chosen to be an \textbf{attention-inducing function} defined as follows.
}

\begin{defi}\label{defi:att-ind}
$A(\|\mathbf{w_i} - \mathbf{w_j}\|^2)$ is an attention-inducing function of $\mathbf{w_i}$ and $\mathbf{w_j}$ if $A:[0,\infty)\rightarrow\R$ is a non-linear function that satisfies the following properties. 
\begin{enumerate}
\item $A$ is increasing and concave on $[0,\infty)$ and $A(0) = 0$;
\item $A$ is continuously differentiable on $(0,\infty)$; and
\item For the derivative $A^\prime$ of $A$, $\lim_{t\rightarrow 0^+} A^\prime(t)$ is finite.
\end{enumerate}
\end{defi}
\nop{According to the above conditions (i) and (ii), }
\nop{
Obviously, an attention-inducing function $D(x,y)$ is an increasing function of the Euclidean distance between $x$ and $y$, and it is continuously differentiable at any $x,y\in\R^n$. 
}

\nop{The class of attention-inducing functions is broad.}

The attention-inducing function $A(\|\mathbf{w_i} - \mathbf{w_j}\|^2)$ measures the difference between $\mathbf{w_i}$ and $\mathbf{w_j}$ in a non-linear manner. 
A typical example of $A(\|\mathbf{w_i} - \mathbf{w_j}\|^2)$ is the negative exponential function $A(\|\mathbf{w_i} - \mathbf{w_j}\|^2) = 1 - e^{-\|\mathbf{w_i} - \mathbf{w_j}\|^2/\sigma}$ with a hyperparameter $\sigma$. Another example is the smoothly clipped absolute deviation function~\cite{fan2001variable}. One more example is the minimax concave penalty function~\cite{zhang2010nearly}.
We adopt the widely-used negative exponential function for our method in this paper.


\nop{
2) the tamed square root function $A(\|w_i - w_j\|^2) = \|w_i - w_j\|^2/(2\sigma)$, where $\|w_i - w_j\|^2\in[0,\sigma^2]$ and $\sigma$ is a user given parameter;
}

\nop{
as well as $\mA(\|w_i - w_j\|^2) = \sqrt{\|w_i - w_j\|^2}-\sigma/2$ when $\|w_i - w_j\|^2>\sigma^2$ and $\sigma>0$ is a parameter that can be chosen by the user.

Also, the smoothly clipped absolute deviation (SCAD) function~\cite{fan2001variable} and the minimax concave penalty (MCP) function~\cite{zhang2010nearly}, which are popular for inducing sparse estimators in high-dimensional statistics, are also examples that satisfy (i) and (ii). 
}

As to be illustrated in the next section, our novel use of the attention-inducing function realizes an attentive message passing mechanism that adaptively facilitates collaborations between clients by iteratively encouraging similar clients to collaborate more with each other. The pairwise collaborations boost the performance in personalized federated learning dramatically.

\section{Federated Attentive Message Passing}
\label{sec:alg}
In this section, we first propose a general method to tackle the optimization problem in Eq.~\eqref{eq:our-form} without considering privacy preservation for clients.
Then, we implement the general method by a personalized federated learning method, \textit{federated attentive message passing} (FedAMP), which collaboratively trains the personalized models of all clients and preserves their data privacy.
Last, we explain why FedAMP can adaptively facilitate collaborations between clients and significantly improve the performance of the personalized models.

\subsection{A General Method}
Denote by $\mF(W)\coloneqq\sum_{i=1}^m F_i(\mathbf{w_i})$ and $\mA(W)\coloneqq\sum_{i<j}^m A(\|\mathbf{w_i} - \mathbf{w_j}\|^2)$ the first and the second terms of $\mG(W)$, respectively. We can rewrite the optimization problem in Eq.~\eqref{eq:our-form} to
\begin{equation}
\label{eq:our-new-form}
\min_{W} \ \left\{\mG(W):= \mF(W) + \lambda \mA(W)\right\}.
\end{equation}

Based on the framework of incremental-type optimization~\cite{bertsekas2011incremental}, we develop a general method to iteratively optimize $\mG(W)$ by alternatively optimizing $\mA(W)$ and $\mF(W)$ until convergence.
In the $k$-th iteration, we first optimize $\mA(W)$  by applying a gradient descent step to compute an intermediate $d$-by-$m$ dimensional matrix
\begin{equation}
\label{eq:grad-descent-intro}
U^k = W^{k-1} - \alpha_k\nabla\mA(W^{k-1}),
\end{equation} 
where $\alpha_k>0$ is the step size of gradient descent, and $W^{k-1}$ denotes the matrix $W$ after the $(k-1)$-th iteration.
Then, we use $U^k$ as the prox-center and apply a proximal point step~\cite{rockafellar1976monotone} to optimize $\mF(W)$ by computing
\begin{equation}
\label{eq:ppa-intro}
W^k = \arg\min_{W} \ \mF(W) + \frac{\lambda}{2\alpha_k}\|W - U^k\|^2.
\end{equation}

This iterative process continues until a preset maximum number of iterations $K$ is reached.
As illustrated later in Section~\ref{sec:analysis}, we analyze the non-asymptotic convergence of the general method, and prove that it converges to an optimal solution when $\mG(W)$ is a convex function, and to a stationary point when $\mG(W)$ is non-convex.

\nop{
this general method non-asymptotically converges \chu{converges to what?} in both the situations when the function $\mG(W)$ is convex and non-convex, respectively.
}

\nop{
to the global optimum point when $\mG(W)$ is convex, and it converges to a local optimum point when $\mG(W)$ is non-convex.
}

\nop{
The non-asymptotic convergence guarantee of this general method is provided in Section~\ref{sec:analysis}.
}

\nop{
In order to conduct personalized federated learning, we regard 
}

\subsection{FedAMP}

The general method introduced above can be easily implemented by merging all clients' private training data together as the training data. To perform personalized federated learning without infringing the data privacy of the clients, we develop FedAMP to implement the optimization steps of the general method in a client-server framework by maintaining a personalized cloud model for each client on a cloud server, and passing weighted model-aggregation messages between personalized models and personalized cloud models.

\nop{
The \textbf{key idea} of FedAMP is to maintain a personalized cloud model for each client on a cloud server, and equivalently conduct the optimization steps of the general method by passing weighted model-aggregation messages between personalized models and personalized cloud models.
}

Following the optimization steps of the general method, FedAMP first optimizes $\mA(W)$ and implements the optimization step in Eq.~\eqref{eq:grad-descent-intro} by computing the $d$-by-$m$ dimensional matrix $U^k$ on the cloud server.

\nop{
Denote by $U^k=[u_1^k, \ldots, u_m^k]$, where the columns $u_1^k, \ldots, u_m^k$ are regarded as the model parameters of the personalized cloud models for each of the $m$ clients, respectively.
}
Let $U^k=[\mathbf{u_1^k}, \ldots, \mathbf{u_m^k}]$, where $\mathbf{u_1^k}, \ldots, \mathbf{u_m^k}$ are the $d$-dimensional columns of $U^k$.
Since $\mA(W):=\sum_{i<j}^m A(\|\mathbf{w_i} - \mathbf{w_j}\|^2)$ and $A(\|\mathbf{w_i} - \mathbf{w_j}\|^2)$ is an attention inducing function, the $i$-th column $\mathbf{u_i^k}$ of matrix $U^k$ computed in Eq.~\eqref{eq:grad-descent-intro} can be rewritten into a linear combination of the model parameter sets $\mathbf{w_1^{k-1}}, \ldots, \mathbf{w_m^{k-1}}$ as follows.
\begin{equation}
\begin{aligned}
\label{eq:grad-step-eq}
\mathbf{u_i^k} =& \left(1 - \alpha_k\sum_{j\neq i}^m A^\prime\left(\|\mathbf{w_i^{k-1}} - \mathbf{w_j^{k-1}}\|^2\right)\right)\cdot \mathbf{w_i^{k-1}} \\ 
&  + \alpha_k \sum_{j\neq i}^m A^\prime\left(\|\mathbf{w_i^{k-1}} - \mathbf{w_j^{k-1}}\|^2\right)\cdot \mathbf{w_j^{k-1}} \\
=& \xi_{i,1}\mathbf{w_1^{k-1}} + \dots + \xi_{i,m}\mathbf{w_m^{k-1}},
\end{aligned}
\end{equation}
where $A^\prime(\|\mathbf{w_i} - \mathbf{w_j}\|^2)$ is the derivative of $A(\|\mathbf{w_i} - \mathbf{w_j}\|^2)$ and $\xi_{i,1}, \ldots, \xi_{i,m}$ are the linear combination weights of the model parameter sets $\mathbf{w_1^{k-1}}, \ldots, \mathbf{w_m^{k-1}}$, respectively.

\nop{
and $\xi_{i,1} + \dots + \xi_{i,m} = 1$ always holds.
}

\nop{, and $u_i^k$ is the model parameters of the $i$-th client's personalized cloud model, that is, a convex combination of the messages received from all the clients including itself}

Often a small value is chosen as the step size $\alpha_k$ of gradient descent so that all the linear combination weights $\xi_{i,1}, \ldots, \xi_{i,m}$ are non-negative. Since $\xi_{i,1} + \cdots + \xi_{i,m} = 1$, $\mathbf{u_i^k}$ is actually a convex combination of the model parameter sets $\mathbf{w_1^{k-1}}, \ldots, \mathbf{w_m^{k-1}}$ of the personalized models of the clients.

\begin{figure}[t]
    \centering
    \includegraphics[width=0.36\textwidth]{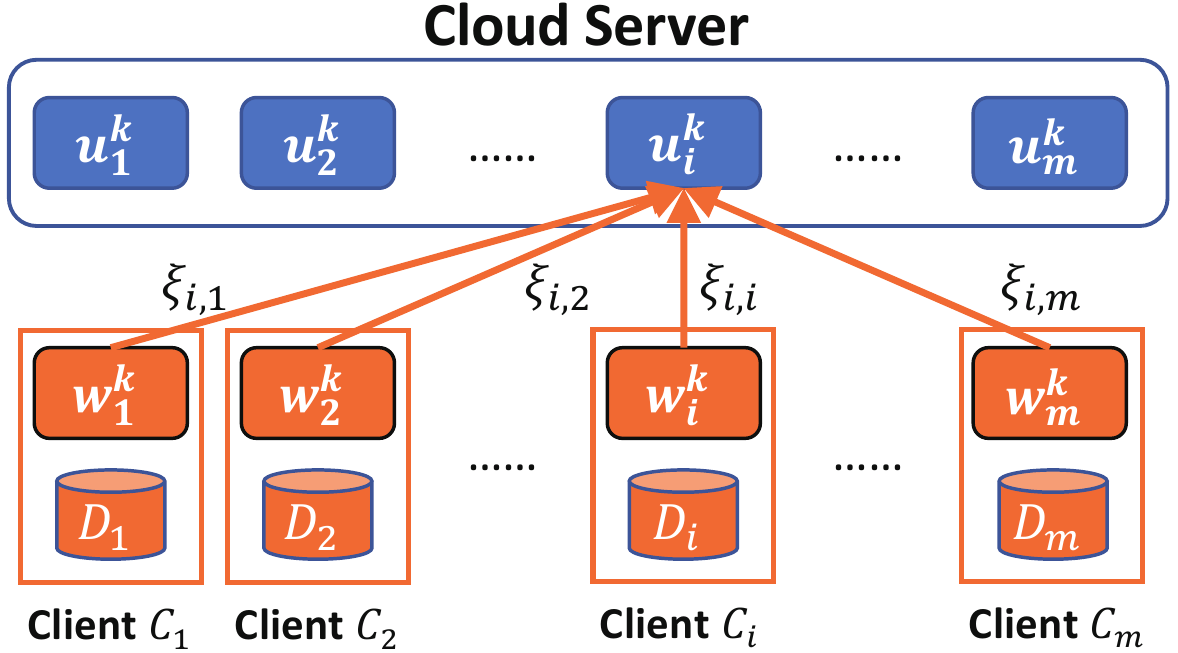} 
    \caption{The message passing mechanism of FedAMP.}
    \label{fig:mapmech}
\end{figure}

As illustrated in Figure~\ref{fig:mapmech}, the convex combination $\mathbf{u_i^k}$ can be modeled a \textit{message passing mechanism} as follows.
We treat $\mathbf{u_i^k}$ as the model parameter set of the \textit{personalized cloud model} of client $C_i$ and also a model aggregation that aggregates $\mathbf{w_1^{k-1}}, \ldots, \mathbf{w_m^{k-1}}$.
Correspondingly, we can treat $\mathbf{w_1^{k-1}}, \ldots, \mathbf{w_m^{k-1}}$ as \textit{model-aggregation messages} that are passed from all clients to client $C_i$ to conduct the model aggregation and produce $\mathbf{u_i^k}$ at the cloud server.

\nop{
we treat $\mathbf{u_i^k}$ as the model parameter sets of the \textit{personalized cloud model} $\mathcal{M}(\mathbf{u_i^k})$ \chu{owned by} the $i$-th client $C_i$, 

We also treat the parameter sets $\mathbf{w_1^{k-1}}, \ldots, \mathbf{w_m^{k-1}}$ as weighted \textit{model-aggregation messages} that are passed from each of the clients to client $C_i$ to conduct the model aggregation on the cloud server.
}

\nop{and
we regard $\mathbf{u_i^k}$ as the aggregation of the model parameter sets $\mathbf{w_1^{k-1}}, \ldots, \mathbf{w_m^{k-1}}$ of the personalized models owned by all clients. 
}


\nop{
by passing weighted model-aggregation messages between personalized models and personalized cloud models. 
}

The above message passing mechanism is the key step for FedAMP to perform inter-client collaboration.
This mechanism solely depends on the model parameter sets $\mathbf{w_1^{k-1}}, \ldots, \mathbf{w_m^{k-1}}$, thus the cloud server can collect $\mathbf{w_1^{k-1}}, \ldots, \mathbf{w_m^{k-1}}$ from the clients and conduct the message passing mechanism to optimize $\mA(W)$ without infringing the data privacy of all the clients.

\nop{
Moreover, 
}
After optimizing $\mA(W)$ on the cloud server, FedAMP then optimizes $\mF(W)$ and implements the optimization step in Eq.~\eqref{eq:ppa-intro} by computing independently columns $\mathbf{w_1^k}, \ldots, \mathbf{w_m^k}$ of $W^k$ for clients $C_1, \ldots, C_m$, respectively.
Recall that $\mathbf{w_i^k}$ is the model parameter set of the personalized model owned by client $C_i$. Following Eq.~\eqref{eq:ppa-intro}, we compute $\mathbf{w_i^k}$ locally on $C_i$ by 
\begin{equation}
\label{eq:local-training}
\mathbf{w_i^{k}} = \arg\min_{\mathbf{w}\in\R^d} F_i(\mathbf{w}) + \frac{\lambda}{2\alpha_{k}}\|\mathbf{w} - \mathbf{u_i^{k}}\|^2,
\end{equation}
Here, we only use the private training data set $D_i$ of client $C_i$ to perform personalized training on model $\mathcal{M}(\mathbf{w_i})$ and, at the same time, consider the inter-client collaboration information carried by the personalized cloud model $\mathcal{M}(\mathbf{u_i^k})$ by requiring $\mathbf{w_i^k}$ and $\mathbf{u_i^k}$ to be close to each other.

\nop{Say something here to clarify that (6) is doing personalization while considering the collaboration with the other clients.}

Since Eq.~\eqref{eq:local-training} only uses $F_i(\mathbf{w})$ and $\mathbf{u_i^k}$, where $F_i(\mathbf{w})$ is determined by the private training data $D_i$ of client $C_i$, $C_i$ can request its own model parameter set $\mathbf{u_i^k}$ from the cloud server and compute $\mathbf{w_i^k}$ locally without exposing its private training data $D_i$ to any other clients or the cloud server. 
Furthermore, since $\mathbf{u_i^k}$ is a convex combination of $\mathbf{w_1^k}, \ldots, \mathbf{w_m^k}$, a client $C_j$ cannot infer the personalized models of any other clients or the private data of any other clients.

\nop{
with $\beta_{k} = \alpha_k/\lambda$, where $\lambda$ is the regularization parameter in Eq.~\eqref{eq:our-form}.
}

\nop{
Following Eq.~\eqref{eq:ppa-intro}, the model parameter $w_i^k$ of the personalized model of client $C_i$ can be computed as
}

\nop{We will discuss why this message passing works in the end of this section.}

\nop{
\begin{equation}\label{eq:cloud-average}
u_i^k = \xi_{i,1}w_1^{k-1} + \dots + \xi_{i,m}w_m^{k-1},
\end{equation}
}

\nop{
$u_i^k\in \mathbb{R}^d$, $i\in\{1, \ldots, m\}$, is the $i$-th column of $U^k$.
}

Algorithm~\ref{alg:FedAMP} summarizes the pseudocode. 
FedAMP implements the optimization steps of the general method in a client-server framework, that is, iteratively optimizing $\mG(W)$ by alternatively optimizing $\mA(W)$ and $\mF(W)$ until a preset maximum number of iterations $K$ is reached.
The non-asymptotic convergence of FedAMP is exactly the same as the general method.

\nop{
\begin{thm}
\label{thm:equivalence}
FedAMP is equivalent to the general method.
\end{thm}

The proof of Theorem~\ref{thm:equivalence} is in \chu{Appendix~\ref{}.}

}

\subsection{Collaboration in FedAMP}

\begin{algorithm}[t]
\DontPrintSemicolon
\SetKwInput{KwInput}{Input}               
\SetKwInput{KwOutput}{Output} 
\KwInput{$m$ clients, each holds a set of private training data and a personalized model to train.}
\KwOutput{The trained model parameter sets $\mathbf{w_1^{K}},\dots, \mathbf{w_m^{K}}$ and $\mathbf{u_1^{K}},\dots, \mathbf{u_m^{K}}$.}

Randomly initialize $\mathbf{w_1^0}, \ldots, \mathbf{w_m^0}$ on the clients.

\For{$k=1,2,\dots,K$}{
Optimize $\mA(W)$: cloud server collects $\mathbf{w_1^{k-1}}, \ldots, \mathbf{w_m^{k-1}}$ from the clients to compute $\mathbf{u_1^{k}}, \dots, \mathbf{u_m^{k}}$ by Eq.~\eqref{eq:grad-step-eq}. \\

Optimize $\mF(W)$: each client $C_i$ requests $\mathbf{u_i^k}$ from the cloud server to compute $\mathbf{w_i^k}$ by Eq.~\eqref{eq:local-training}.
\nop{
\For{$i=1,2,\ldots,m$}{
    Client $C_i$ requests $u_i^k$ from the cloud server to compute $w_i^k$ by Eq.~\eqref{eq:local-training}.\\
}}
}
\caption{FedAMP}
\label{alg:FedAMP}
\end{algorithm}

\nop{
Now we discuss why FedAMP realizes the attentive collaboration between clients.
}

\nop{
\textbf{First of all}, since {FedAMP} never transfers the private data of any client, the data privacy of all clients is kept intact during the entire training process.
}

\nop{
\textbf{Second}, {FedAMP} is a message passing approach. We can observe from Eq.~\eqref{eq:grad-step-eq} that the cloud model $u^k_i$ can be rewritten in the following form
\begin{equation}\label{eq:cloud-average}
u_i^k = \xi_{i,1}w_1^{k-1} + \dots + \xi_{i,m}w_m^{k-1}
\end{equation}
where $\xi_{i,1} + \dots + \xi_{i,m} = 1$, and $(\xi_{i,1}, \ldots, \xi_{i,m})$ are linear combination weights for the client model parameters $(w_1^{k-1}, \ldots, w_m^{k-1})$.
Obviously, the client model parameters $(w_1^{k-1}, \ldots, w_m^{k-1})$ can be viewed as the \textbf{messages} that are \textbf{passed} from all clients to the $i$-th client, and the model parameters $u_i^k$ of client $i$'s personalized cloud model is a weighted aggregation of the messages received from all clients.
As a result, {FedAMP} is a message passing approach that conducts inter-client collaboration by passing messages between local models and personalized cloud models. 
}

\nop{===============}

FedAMP adaptively facilitates collaborations between similar clients, since the attentive message passing mechanism iteratively encourages similar clients to collaborate more with each other during the personalized federated learning process.

\nop{
why the message passing process in Eq.~\eqref{eq:grad-step-eq} is attentive, that is, it encourages similar clients to collaborate more with each other, and further conclude the potential of this attentive message passing process in adaptively discovering the underlying collaboration relationships between clients.
}

To analyze the attentive message passing mechanism of FedAMP, 
we revisit the weights $\xi_{i,1}, \ldots, \xi_{i,m}$ of the convex combination in Eq.~\eqref{eq:grad-step-eq}, where the weight
\begin{equation}
\label{eq:attentive-mechanism}
    \xi_{i,j}=\alpha_k A^\prime\left(\|\mathbf{w_i^{k-1}} - \mathbf{w_j^{k-1}}\|^2\right), (i\neq j)
\end{equation}
is the contribution of message $\mathbf{w_j^{k-1}}$ sent from client $C_j$ to the aggregated model parameter set $\mathbf{u_i^{k}}$ of the personalized cloud model owned by client $C_i$.
$\xi_{i,i} = 1 - \sum_{j\neq i}^m \xi_{i,j}$
is simply a self-attention weight that specifies the proportion of the model parameter set $\mathbf{w_i^{k-1}}$ of client $C_i$'s personalized model in its own personalized cloud model.

Due to Definition \ref{defi:att-ind}, $A$ is an increasing and concave function on $[0,\infty)$. Thus, the derivative $A^\prime$ of $A$ is a non-negative and non-increasing function on $(0,\infty)$.
Therefore, function $A^\prime(\|\mathbf{w_i^{k-1}} - \mathbf{w_j^{k-1}}\|^2)$ is a \textit{similarity function} that measures the similarity between $\mathbf{w_i^{k-1}}$ and $\mathbf{w_j^{k-1}}$, such that their similarity is high if they have a small Euclidean distance.

From Eq.~\eqref{eq:attentive-mechanism},
if the model parameters $\mathbf{w_i^{k-1}}$ and $\mathbf{w_j^{k-1}}$ are similar with each other, they contribute more to the model parameters $\mathbf{u_j^k}$ and $\mathbf{u_i^k}$ of clients $C_j$ and $C_i$, respectively.
This further makes $\mathbf{u_i^k}$ and $\mathbf{u_j^k}$ more similar to each other.
Since the optimization step in Eq.~\eqref{eq:local-training} forces $\mathbf{w_i^k}$ and $\mathbf{w_j^k}$ to be close to $\mathbf{u_i^k}$ and $\mathbf{u_j^k}$, respectively, $\mathbf{w_i^k}$ and $\mathbf{w_j^k}$ are more similar to each other as well.
 
In summary, FedAMP builds a positive feedback loop that iteratively encourages clients with similar model parameters to have stronger collaborations, and adaptively and implicitly groups similar clients together to conduct more effective collaborations.

\section{Convergence Analysis of FedAMP}\label{sec:analysis}
In this section, we analyze the convergence of FedAMP when $\mG$ is convex or non-convex under suitable conditions. 
To begin with, similar to the analysis of many incremental and stochastic optimization algorithms~\cite{bertsekas2011incremental,nemirovski2009robust}, we make the following assumption.
\begin{assumption}\label{ass:bound}
There exists a constant $B>0$ such that $\max\{ \|Y\|: Y\in \partial \mF(W^k) \} \leq B$ and $\|\nabla \mA(W^k)\|\leq B/\lambda$ hold for every  $k \geq 0$, where $\partial \mF$ is the subdifferential of $\mF$ and $\|\cdot\|$ is the Frobenius norm. 
\nop{
\begin{equation}\nonumber
\max\{ \|Y\|: Y\in \partial \mF(w_1^k,\dots,w_m^k) \} \leq B
\end{equation}
and
\begin{equation}\nonumber
\|\nabla \mD(w_1^k,\dots,w_m^k)\|\leq B/\lambda.
\end{equation}
}
\end{assumption}

For our problem in Eq.~\eqref{eq:our-form}, Assumption~\ref{ass:bound} naturally holds if both $\mF(W)$ and $\mA(W)$ are locally Lipschitz continuous and $\|W^k\|$ is bounded by a constant for all $k\geq 0$. 

\nop{
\chu{bounded by what? a large enough finite constant?}.
}

Now, we provide the guarantee on convergence for FedAMP when both $\mF(W)$ and $\mA(W)$ are convex functions.
\nop{
When both $\mF$ and $\mD$ are convex, we have the following guarantee of {FedAMP}.
}
\begin{thm}
\label{thm:analysis-convex}
Under Assumption \ref{ass:bound} and assuming functions $\mF(W)$ and $\mA(W)$ in Eq.~\eqref{eq:our-form} are convex, if $\alpha_{1} = \cdots = \alpha_K = \lambda/\sqrt{K}$ for some $K\geq 0$, then the sequence $W^0, \ldots, W^K$ generated by Algorithm \ref{alg:FedAMP} satisfies
\begin{equation}\nonumber
\min_{0\leq k\leq K} \mG(W^k) \leq \mG^* + \frac{\|W^0 - W^*\|^2 + 5B^2}{\sqrt{K}},
\end{equation}
where $W^*$ is an optimal solution of Eq.~\eqref{eq:our-form} and $\mG^*= \mG(W^*)$. Moreover, if $\alpha_k$ satisfies $\sum_{k=1}^\infty\alpha_k = \infty$ and $\sum_{k=1}^\infty\alpha_k^2<\infty$, then 
\begin{equation}
\nonumber
\liminf_{k\rightarrow\infty} \mG(W^k) = \mG^*.
\end{equation}
\end{thm}

Theorem~\ref{thm:analysis-convex} implies that for any $\epsilon>0$, FedAMP needs at most $\mathcal{O}(\epsilon^{-2})$ iterations to find an $\epsilon$-optimal solution $\widetilde{W}$ of Eq.~\eqref{eq:our-form} such that $\mG(\widetilde{W}) - \mG^* \leq \epsilon$. It also establishes the global convergence of FedAMP to an optimal solution of Eq.~\eqref{eq:our-form} when $\mG$ is convex. The proof of Theorem \ref{thm:analysis-convex} is provided in Appendix~A~\cite{huang2020personalized}. 

\nop{\chu{To Zirui: Please write the remark of Theorem~\ref{thm:analysis-convex} in here, and also build the logical connection between Theorem~\ref{thm:analysis-convex} and our previous claim that ``$\mG(W)$ converges to an optimal point when it is convex''. Thanks!}}

\nop{
Recall that $\partial \mF(W) = \{\nabla\mF(W)\}$ if $\mF$ is differentiable at $W$.}

Next, we provide the convergence guarantee of {FedAMP} when $\mG(W)$ is a smooth and non-convex function. 

\begin{thm}
\label{thm:analysis-ncvx}
Under Assumption~\ref{ass:bound} and assuming functions $\mF(W)$ and $\mA(W)$ in Eq.~\eqref{eq:our-form} are continuously differentiable and the gradients $\nabla\mF(W)$ and $\nabla \mA(W)$ are Lipschitz continuous with modulus $L$, 
if $\alpha_{1} = \cdots = \alpha_K = \lambda/\sqrt{K}$, then the sequence $W^0, \ldots, W^K$ generated by Algorithm \ref{alg:FedAMP} satisfies
\begin{equation}\nonumber
\begin{aligned}
& \min_{0\leq k\leq K} \|\nabla \mG(W^k)\|^2 \\ 
& \quad \leq \frac{18(\mG(W^0) - \mG^* + 20LB^2)}{\sqrt{K}} + \mathcal{O}\left(\frac{1}{K}\right)
\end{aligned}
\end{equation}
where $W^*$ and $\mG^*$ are the same as in Theorem \ref{thm:analysis-convex}. Moreover, if $\alpha_k$ satisfies $\sum_{k=1}^\infty\alpha_k = \infty$ and $\sum_{k=1}^\infty\alpha_k^2<\infty$, then
\begin{equation}
\nonumber
\liminf_{k\rightarrow\infty} \|\nabla\mG(W^k)\| = 0.
\end{equation} 
\end{thm}

Theorem \ref{thm:analysis-ncvx} implies that for any $\epsilon>0$, FedAMP needs at most $\mathcal{O}(\epsilon^{-4})$ iterations to find an $\epsilon$-approximate stationary point $\widetilde{W}$ of Eq.~\eqref{eq:our-form} such that $\|\nabla\mG(\widetilde{W})\|\leq \epsilon$. It also establishes the global convergence of FedAMP to a stationary point of Eq.~\eqref{eq:our-form} when $\mG$ is smooth and non-convex. The proof of Theorem~\ref{thm:analysis-ncvx} is in Appendix~B~\cite{huang2020personalized}.

\nop{\chu{To Zirui: Please write the remark of Theorem~\ref{thm:analysis-ncvx} in here, and also build the logical connection between Theorem~\ref{thm:analysis-ncvx} and our previous claim that ``$\mG(W)$ converges to a stationary point when it is non-convex''. Thanks!}}


\section{HeurFedAMP: Heuristic Improvement of FedAMP on Deep Neural Networks}
\label{sec:instances}

In this section, we tackle the challenge in the message passing mechanism when deep neural networks are used by clients, and propose a heuristic improvement of FedAMP.


\nop{ the attentive message passing mechanism of FedAMP iteratively encourages the clients with similar personalized models to collaborate more by passing model aggregation messages with larger weights to each other. }

\nop{
The message weights $\xi_{i,1}, \ldots, \xi_{i,m}$ are determined by the similarity function $\mA^\prime$ that evaluates the similarity between the personalized models of different clients.
A good similarity function that accurately measures the model similarity will produce a set of high quality message weights $\xi_{i,1}, \ldots, \xi_{i,m}$. This will boost the effectiveness of the attentive message passing mechanism to achieve good personalized federated learning performance.
}

As illustrated in Section~\ref{sec:alg}, the effectiveness of the attentive message passing mechanism of FedAMP
largely depends on the weights $\xi_{i,1}, \ldots, \xi_{i,m}$ of the model aggregation messages. These message weights are determined by the similarity function $A^\prime(\|\mathbf{w_i} - \mathbf{w_j}\|^2)$ that
measures the similarity between the model parameter sets $\mathbf{w_i}$ and $\mathbf{w_j}$ based on their Euclidean distance $\|\mathbf{w_i} - \mathbf{w_j}\|$.

\nop{
evaluates the similarities between the personalized models of different clients.
}
\nop{
the weight $\xi_{i,j}$ of the model aggregation message passed from client $C_j$ to client $C_i$ is determined by the similarity between their personalized models $\mathcal{M}(w_i)$ and $\mathcal{M}(w_j)$. }

\nop{
For FedAMP, the similarity between the personalized models $\mathcal{M}(w_i)$ and $\mathcal{M}(w_j)$ directly determines the weight $\xi_{i,j}$ of the model aggregation message passed from client $C_j$ to client $C_i$.
This similarity is evaluated by the similarity function $\mA^\prime(\| w_i - w_j\|^2)$ that measures the similarity between $w_i$ and $w_j$ based on their Euclidean distance $\|w_i - w_j\|$. 
}

When the dimensionalities of $\mathbf{w_i}$ and $\mathbf{w_j}$ are small, Euclidean distance is a good measurement to evaluate their difference. In this case, the similarity function $A^\prime(\|\mathbf{w_i} - \mathbf{w_j}\|^2)$ works well in evaluating the similarity between $\mathbf{w_i}$ and $\mathbf{w_j}$. 
However, when clients adopt deep neural networks as their personalized models, each personalized model involves a large number of parameters, which means the dimensionalities of both $\mathbf{w_i}$ and $\mathbf{w_j}$ are high.
In this case, Euclidean distance may not be effective in evaluating the difference between $\mathbf{w_i}$ and $\mathbf{w_j}$ anymore due to the curse of dimensionality~\cite{verleysen2005curse}. Consequently, the message weights produced by $A^\prime(\|\mathbf{w_i} - \mathbf{w_j}\|^2)$ may not be an effective attentive message passing mechanism.  Thus, we need a better way to produce the message weights instead of using $A^\prime(\|\mathbf{w_i} - \mathbf{w_j}\|^2)$.

To tackle the challenge, we propose HeurFedAMP, a heuristic revision of FedAMP when clients use deep neural networks.
The key idea of HeurFedAMP is to heuristically compute the message weights in a different way that works well with the high-dimensional model parameters of deep neural networks. 
Specifically, HeurFedAMP follows the optimization steps of FedAMP exactly, except that, when computing message weights $\xi_{i,1}, \ldots, \xi_{i,m}$ in the $k$-th iteration, HeurFedAMP first treats weight $\xi_{i,i}$ as a self-attention hyper-parameter that controls the proportion of the message $\mathbf{w_i^{k-1}}$ sent from client $C_i$ to its own personalized cloud model, and then computes the weight of the message passed from a client $C_j$ to client $C_i$ by
\begin{equation}
\label{eq:heurweights}
\xi_{i,j} = \frac{e^{\sigma\cos(\mathbf{w_i^{k-1}},\mathbf{w_j^{k-1}})}}
{\sum_{h\neq i}^m e^{\sigma\cos(\mathbf{w_i^{k-1}},\mathbf{w_h^{k-1}})}}\cdot (1-\xi_{i,i}),
\end{equation}
where $\sigma$ is a scaling hyper-parameter and $\cos(\mathbf{w_i^{k-1}},\mathbf{w_j^{k-1}})$ is the cosine similarity between $\mathbf{w_i^{k-1}}$ and $\mathbf{w_j^{k-1}}$.

All the weights $\xi_{i,1}, \ldots, \xi_{i,m}$ computed by HeurFedAMP are non-negative and sum to 1. Applying the weights computed by HeurFedAMP to Eq.~\eqref{eq:grad-step-eq}, the model parameter set $\mathbf{u_i^k}$ of the personalized cloud model of client $C_i$ is still a convex combination of all the messages that it receives.

Furthermore, according to from Eq.~\eqref{eq:heurweights}, 
if the model parameter sets $\mathbf{w_i^{k-1}}$ and $\mathbf{w_j^{k-1}}$ of two clients have a large cosine similarity $\cos(\mathbf{w_i^{k-1}}, \mathbf{w_j^{k-1}})$, their messages have large weights and contribute more to the personalized cloud models of each other. In other words, HeurFedAMP builds a positive feedback loop similar to that of FedAMP to realize the attentive message passing mechanism.

As to be demonstrated in Section~\ref{sec:exp}, HeurFedAMP improves the performance of FedAMP when clients adopt deep neural networks as personalized models, because cosine similarity is well-known to be more robust in evaluating similarity between high dimensional model parameters than Euclidean distance.

\section{Experiments}
\label{sec:exp}

In this section, we evaluate the performance of {FedAMP} and {HeurFedAMP} and compare them with the state-of-the-art personalized federated learning algorithms, including SCAFFOLD~\cite{karimireddy2019scaffold}, APFL~\cite{deng2020adaptive}, FedAvg-FT and FedProx-FT~\cite{wang2019federated}.
FedAvg-FT and FedProx-FT are two local fine-tuning methods~\cite{wang2019federated} that obtain personalized models by fine-tuning the global models produced by the classic global federated learning methods
FedAvg~\cite{mcmahan2016communication} and FedProx~\cite{li2018federated}, respectively.
To make our experiments more comprehensive, we also report the performance of FedAvg, FedProx and a naive separate training method named \emph{Separate} that independently trains the personalized model of each client without collaboration between clients. 

The performance of all the methods is evaluated by the \emph{best mean testing accuracy} (BMTA) in percentage, where the \emph{mean testing accuracy} is the average of the testing accuracies on all clients, and BMTA is the highest mean testing accuracy achieved by a method during all the communication rounds of training.

All the methods are implemented in PyTorch 1.3 running on Dell Alienware with Intel(R) Core(TM) i9-9980XE CPU, 128G memory, NVIDIA 1080Ti, and Ubuntu 16.04.

\subsection{Settings of Data Sets}
\label{subsec:datasettings}
We use four public benchmark data sets, MNIST~\cite{lecun2010mnist}, FMNIST (Fashion-MNIST)~\cite{xiao2017fashion}, EMNIST (Extended-MNIST)~\cite{cohen2017emnist} and CIFAR100~\cite{krizhevsky2009learning}. 

For each of the data sets, we apply three different data settings: 
1) an IID data setting~\cite{mcmahan2016communication} that uniformly distributes data across different clients; 
2) a pathological non-IID data setting~\cite{mcmahan2016communication} that partitions the data set in a non-IID manner such that each client contains two classes of samples and there is no group-wise similarities between the private data of clients; 
and 
3) a practical non-IID data setting that first partitions clients into groups, and then assigns data samples to clients in such a way that the clients in the same group have similar data distributions, the clients in different groups have different data distributions, every client has data from all classes, and the number of samples per client is different for different groups. 

Comparing with the pathological non-IID data setting, the practical non-IID data setting is closer to reality, since in practice each company participating in a personalized federated learning process often has data from most of the classes, and it is common that a subgroup of companies may have similar data distributions that are different from the data owned by companies outside the subgroup.

\nop{
the same data distribution to the clients in the same group while assigning different data distribution to the clients in the different groups, where each client has data from all classes and the number of samples is different for clients in different groups.
}

\nop{
with unbalanced number of samples 

The practical non-IID data setting is a more common scenario in practice, because most real world companies that participate in a personalized federated learning process will have data from all classes and the number of samples is unbalanced across different groups.
}

\nop{
process will have data from all classes and the number of samples is unbalanced across different groups.

}

\nop{
Based on the description above, we can see that comparing with the pathological non-IID setting~\cite{mcmahan2016communication}, every client in the practical non-IID setting has data from all classes and the number of samples is unbalanced across different groups, which is a more common scenario in practice.
}

\nop{
In the IID setting, the data is uniformly distributed across different clients as described in~\cite{mcmahan2016communication}. In the pathological non-IID setting, we follow the steps provided in~\cite{mcmahan2016communication}, where they partition the data set based on labels and on each client, samples are drawn only from two classes. 
}

\nop{
In the practical non-IID setting, we first create groups among the clients and then we assign same data distribution to the clients in the same group while assign different data distribution to the clients in the different groups.
}

Let us take EMNIST as an example to show how we apply the practical non-IID data setting. 
First, we set up 62 clients numbered as clients $0, 1, \ldots, 61$ and divide them into three groups.
Then, we assign the samples to the clients such that 80\% of the data of every client are uniformly sampled from a set of dominating classes, and 20\% of the data are uniformly sampled from the rest of the classes.
Specifically, the first group consists of clients 0-9, where each client has 1000 training samples from the dominating classes with digit labels from `0' to `9'.
The second group consists of clients 10-35, where each client has 700 training samples from the dominating classes of upper-case letters from `A' to `Z'. The third group consists of clients 36-61, where each client has 400 training samples from the dominating classes of lower-case letters from `a' to `z'.
Every client has 100 testing samples with the same distribution as its training data.

\nop{
In addition, each client has 100 testing samples. On each client, the dominated classes uniformly have 80\% of data while the non-dominated classes uniformly have the rest 20\% data.
}

Limited by space, we only report the most important experimental results in the rest of this section. 
Please see Appendix~C~\cite{huang2020personalized} for the details of the practical non-IID data setting on MNIST, FMNIST and CIFAR100, the implementation details and the hyperparameter settings of all the methods, and also more extensive results about the convergence and robustness of the proposed methods.

\subsection{Results on the IID Data Setting}\label{subsec:expres-1}
Table~\ref{tab:result_IID} shows the BMTA of all methods being compared under the IID data setting.
The performance of Separate is a good baseline to indicate the needs of collaboration on classifying the data sets, since Separate does not conduct collaboration at all.
Separate achieves a performance comparable with all the other methods on the easy data set MNIST. However, on the more challenging data sets FMNIST, EMNIST and CIFAR100, the performance of Separate is significantly behind that of the others due to the lack of collaborations between clients.

The global federated learning methods FedAvg and FedProx achieve the best performance most of the time on IID data, because the clients are similar to each other and the global model fits every client well. Differentiating pairwise collaborations between different clients are not needed on IID data.
APFL achieves a performance comparable with FedAvg and FedProx on all data sets, because it degenerates to FedAvg under the IID data setting~\cite{deng2020adaptive}. For this reason, under the IID data setting, we consider APFL a global federated learning method instead of a personalized federated learning method.

The personalized federated learning methods FedAvg-FT, FedProx-FT and SCAFFOLD do not perform as well as FedAvg and FedProx under the IID data setting.
Although they achieve a performance comparable to FedAvg and FedProx on MNIST, their performances on the more challenging data sets FMNIST, EMNIST and CIFAR100 are clearly inferior to FedAvg and FedProx.  The local fine-tuning steps of FedAvg-FT and FedProx-FT are prone to over-fitting, and the rigid customization on the gradient updates of SCAFFOLD limits its flexibility to fit IID data well.

FedAMP and HeurFedAMP perform much better than FedAvg-FT, FedProx-FT and SCAFFOLD under the IID data setting. The personalized models of clients are similar to each other under the IID data setting, thus the attentive message passing mechanism assigns comparable weights to all messages, which accomplishes a global collaboration among all clients similar to that of FedAvg and FedAMP in effect.
FedAMP and HeurFedAMP achieve the best performance among all the personalized federated learning methods on all data sets, and also perform comparably well as FedAvg and FedProx on MNIST, FMNIST and EMNIST.

\begin{table}[t]
\setlength{\tabcolsep}{2pt}
\centering
\begin{tabular}{l|cccc} 
 \toprule
Methods & MNIST & FMNIST & EMNIST & CIFAR100 \\ 
 \midrule
 Separate & 99.27	& 81.66 &	54.41 &	 9.82 \\
 \hline
 FedAvg  & 99.31 &	91.94	& 74.38 &	49.59
 \\
 FedProx & 98.81 &	90.19 & 73.14	& 46.50
 \\
 \hline
 FedAvg-FT & 98.98 & 90.17 &	70.53	& 35.07
 \\
  FedProx-FT &  98.72 & 89.02 &	69.49	& 40.77	\\
  SCAFFOLD & 98.89 & 89.04 & 72.51 & 43.06 \\
  APFL & 98.93 & 91.03 & 73.95 & 49.02 \\
  \hline
 {FedAMP} & 99.22 & 92.05 & 74.07 & 45.68\\ 
 {HeurFedAMP} & 99.28	& 91.80 & 74.07 &  45.88\\
 \bottomrule
\end{tabular}
\caption{BMTA for the IID data setting.}
\label{tab:result_IID}
\end{table}

\nop{
In Table~\ref{tab:result_NonIIDNonGrouping}, we summarize the performance of {FedAMP}, {HeurFedAMP} and baselines in terms of the best mean testing accuracy for the pathological non-IID data setting. It is easy to observe that the separate training in general performs very well in this data setting. In addition, for all the data sets, the performance of almost all the local customization methods, i.e., FedAvg-FT, FedProx-FT, SCAFFOLD, APFL, {FedAMP}, and {HeurFedAMP} is similar to the performance of the separate training. 
The reason for these observations is that in this data setting, each client only contains 2 classes of samples. This makes the training much simpler when comparing with the IID data setting, where all clients have samples from all classes. Thus, the pathological non-IID data setting is not a practical non-IID data distribution for evaluating personalized federated learning algorithms. 

}

\begin{table}[t]
\setlength{\tabcolsep}{2pt}
\centering
\begin{tabular}{l|cccc} 
 \toprule
Methods & MNIST & FMNIST & EMNIST & CIFAR100 \\ 
 \midrule
 Separate & 98.73 & 97.67 &	99.15 &	  92.67\\
 \hline
 FedAvg  & 98.39 & 77.88 & 19.44 &  2.70\\ 
 FedProx & 97.15 & 83.80 & 48.81 &  2.81\\ 
 \hline
 FedAvg-FT & 99.66 & 98.07 & 99.24 &  95.00 \\
 FedProx-FT & 99.63 & 98.00 & 99.27 & 94.36 \\
 SCAFFOLD & 99.34 & 94.58 & 98.75 & 2.04\\
 APFL & 98.24 & 97.44 & 98.90 & 52.11\\
 \hline
 {FedAMP} & 99.53 & 97.95 &	99.27 & 94.87\\ 
 {HeurFedAMP} & 99.38 & 98.17 & 99.26 &  94.74\\
 \bottomrule
\end{tabular}
\caption{BMTA for the pathological non-IID data setting.}
\label{tab:result_NonIIDNonGrouping}
\end{table}

\subsection{Results on the Pathological Non-IID Data Setting}\label{subsec:expres-2}
Table~\ref{tab:result_NonIIDNonGrouping} shows the BMTA of all the methods under the pathological non-IID data setting. 
This data setting is pathological because each client contains only two classes of samples, which largely simplifies the classification task on every client~\cite{mcmahan2016communication}.
The simplicity of client tasks is clearly indicated by the high performance of Separate on all the data sets.

However, the pathological non-IID data setting is not easy for the global federated learning methods.
The performance of FedAvg and FedProx degenerates a lot on FMNIST and EMNIST, because taking the global aggregation of all personalized models trained on the non-IID data of different clients introduces significant unstableness to the gradient-based optimization process~\cite{zhang2020fedpd}.

On the most challenging CIFAR100 data set,
the unstableness catastrophically destroys the performance of the global models produced by FedAvg and FedProx, and also significantly damages the performance of SCAFFOLD and APFL because the global models are destroyed
such that the customized gradient updates of SCAFFOLD and the model mixtures conducted by APFL can hardly tune it up. 

The other personalized federated learning methods FedAvg-FT, FedProx-FT, FedAMP and HeurFedAMP achieve comparably good performance on all data sets. 
FedAvg-FT and FedProx-FT achieve good performance by taking many fine-tuning steps to tune the poor global models back to normal.
The good performance of FedAMP and HeurFedAMP is achieved by adaptively facilitating pair-wise collaborations between clients without using a single global model. Since the personalized cloud models of 
FedAMP and HeurFedAMP only aggregate similar personalized models of clients, they stably converge without suffering from the unstableness caused by the global aggregation of different personalized models.

\begin{table}[t]
\setlength{\tabcolsep}{2pt}
\centering
\begin{tabular}{l|ccccc} 
 \toprule
Methods & MNIST & FMNIST & EMNIST & CIFAR100 \\
 \midrule
 Separate & 86.30	& 86.73 &	61.78 &	39.99 \\
 \hline
 FedAvg  & 81.82 &	79.50	& 72.27 &	35.21
 \\
 FedProx & 81.46 &	78.71 &	70.55	& 37.31
 \\
 \hline
 FedAvg-FT & 91.79 &	89.73 &	78.93	& 49.00\\
 FedProx-FT &  94.10 & 87.51 &	77.31	&	50.24\\
 SCAFFOLD & \textbf{98.50} & 40.20 & 77.98 & 21.29\\
 APFL & 85.05 & 84.08 & 59.07 & 16.45\\
 \hline
 {FedAMP} & 97.59 & 90.97 & 81.22 & 53.04\\ 
 {HeurFedAMP} & 97.36	& \textbf{91.37} & \textbf{81.47} & \textbf{53.27} \\
 \bottomrule
\end{tabular}
\caption{BMTA for the practical non-IID data setting.}
\label{tab:result}
\end{table}

\subsection{Results on the Practical Non-IID Data Setting}\label{subsec:expres}

Table~\ref{tab:result} evaluates all methods in BMTA under the practical non-IID data setting. FedAMP and HeurFedAMP perform comparably well as SCAFFOLD on MNIST, and they significantly outperform all other methods on FMNIST, EMNIST and CIFAR100.

\begin{figure}[t]
    \centering
    \begin{subfigure}[b]{0.46\textwidth}
        \centering
        \includegraphics[width=\textwidth]{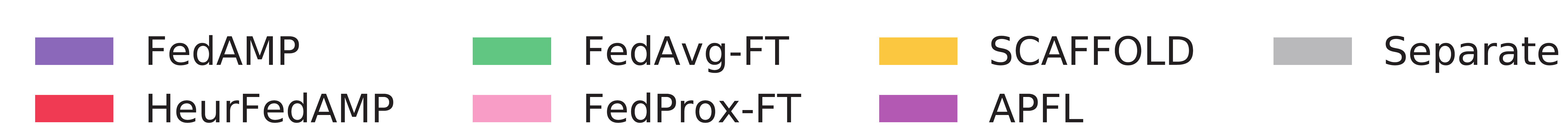}
    \end{subfigure}
    \begin{subfigure}[b]{0.23\textwidth}
        \centering
        \includegraphics[width=\textwidth]{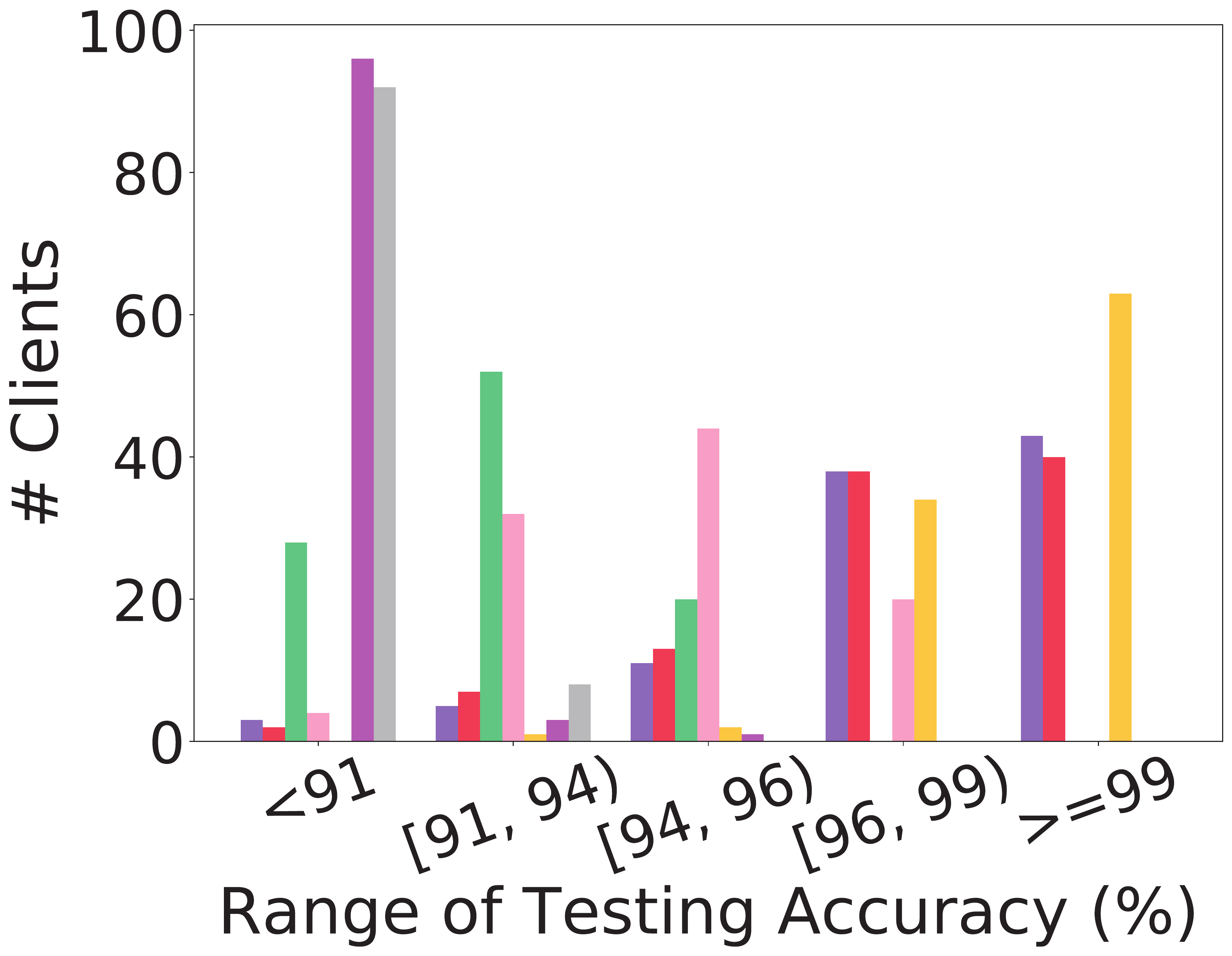} 
        \caption{MNIST}
        \label{subfig:MNIST}
    \end{subfigure}
    \begin{subfigure}[b]{0.23\textwidth}
        \centering
        \includegraphics[width=\textwidth]{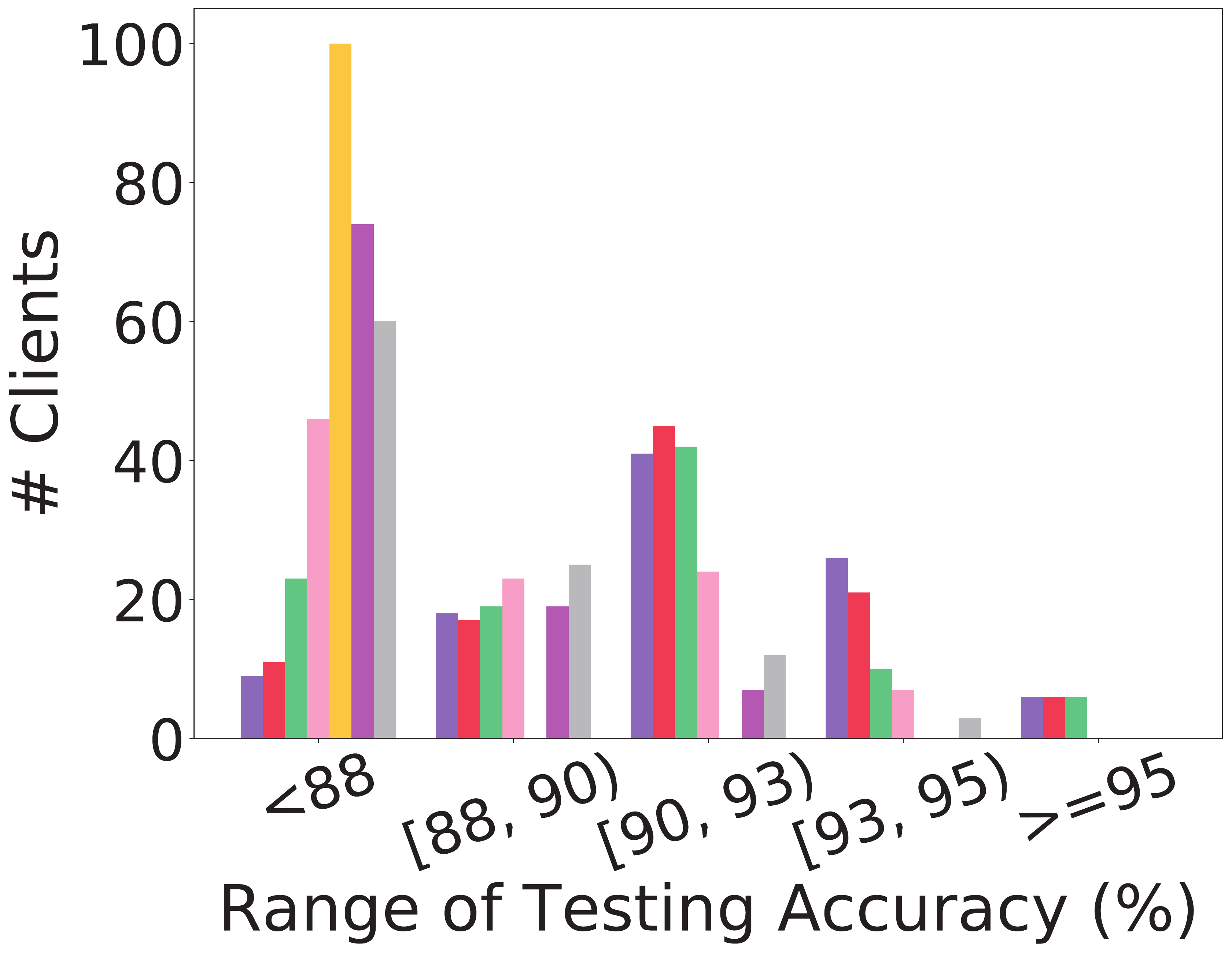}
        \caption{FMNIST}
        \label{subfig:FMNIST}
    \end{subfigure}
    \begin{subfigure}[b]{0.23\textwidth}
        \centering
        \includegraphics[width=\textwidth]{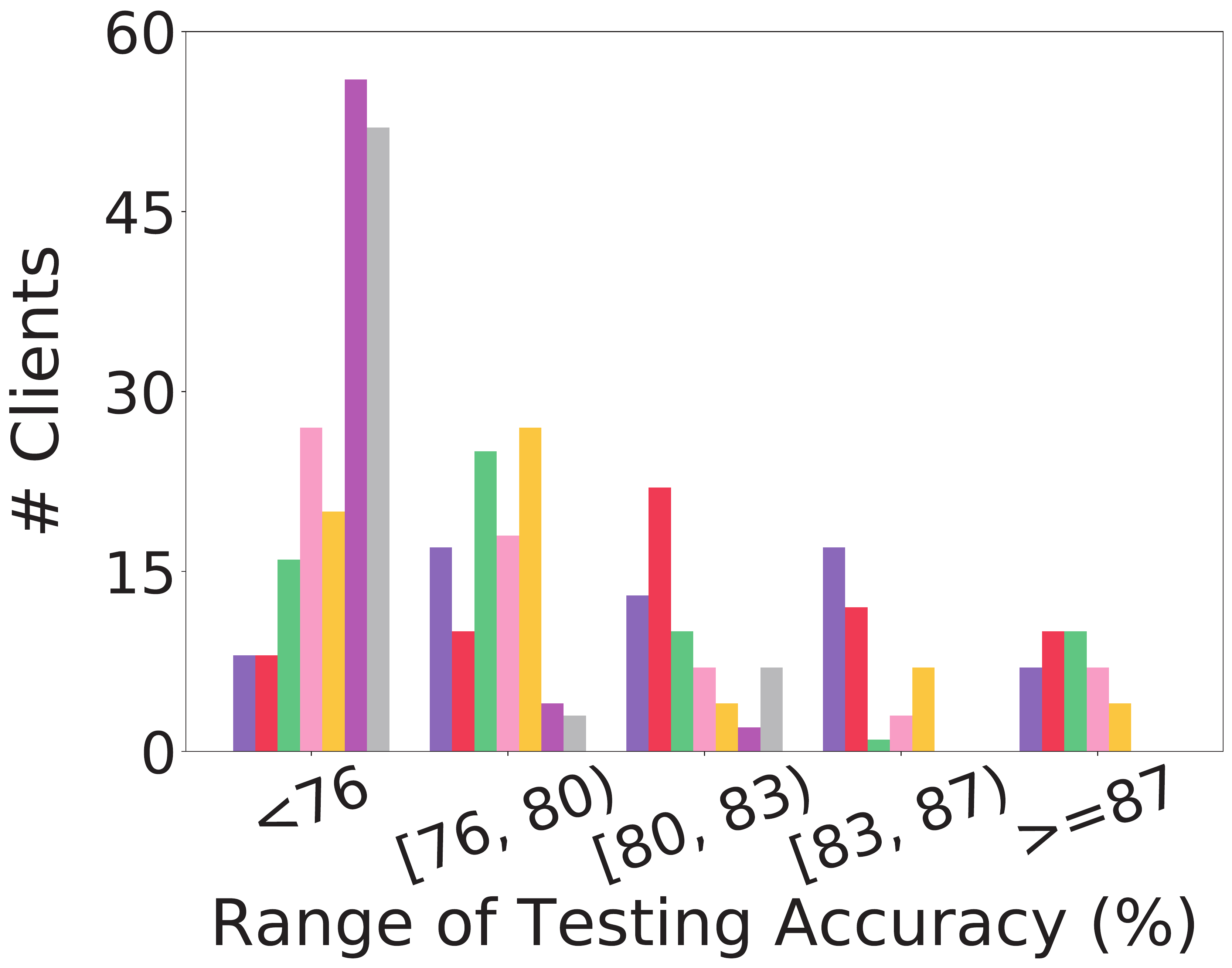} 
        \caption{EMNIST}
        \label{subfig:EMNIST}
    \end{subfigure}
    \begin{subfigure}[b]{0.23\textwidth}
        \centering
        \includegraphics[width=\textwidth]{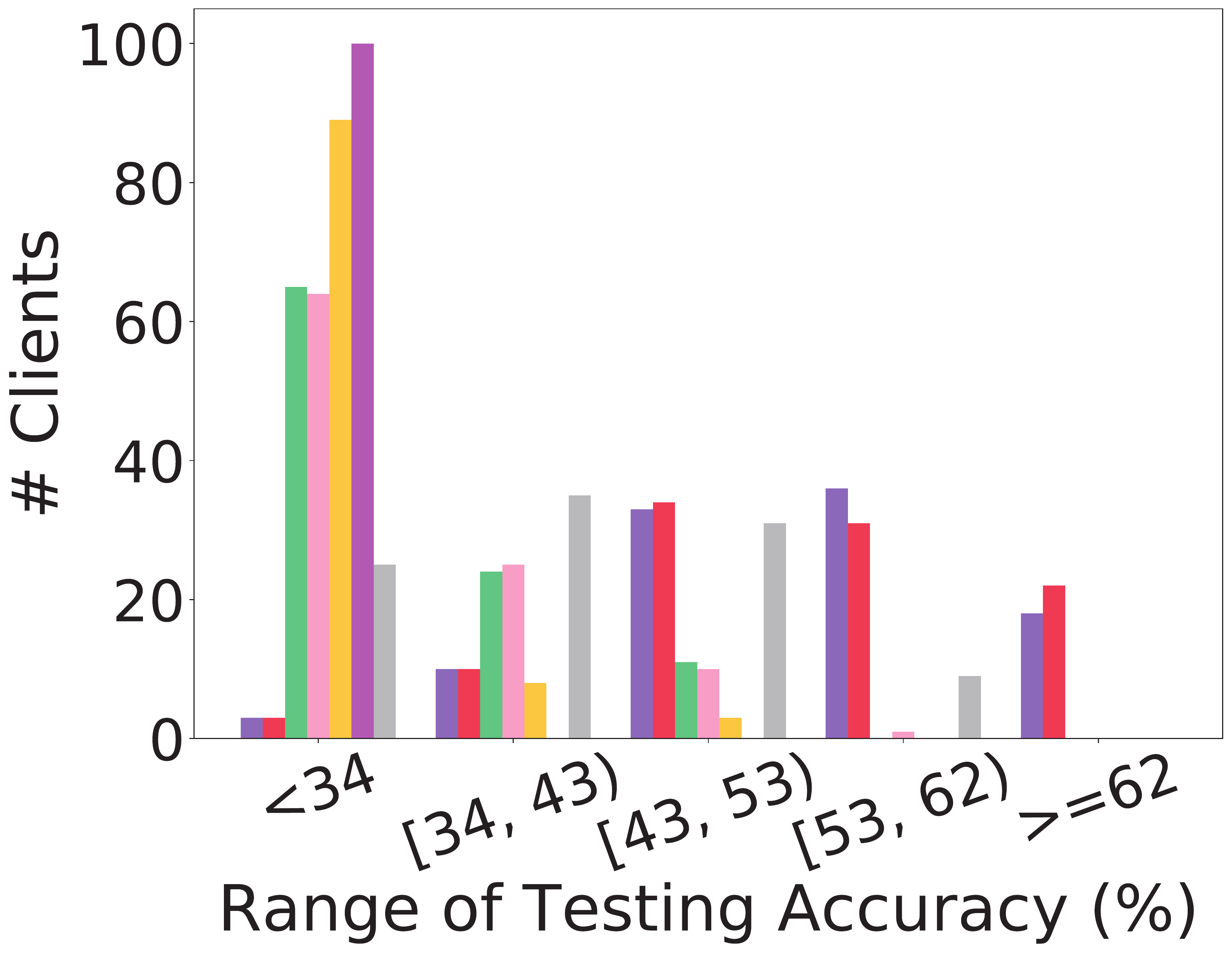}
        \caption{CIFAR100}
        \label{subfig:cifar}
    \end{subfigure}
    \caption{The distribution of the testing accuracy of all clients under the practical non-IID data setting.}
    \label{fig:distribution}
\end{figure}

To evaluate the personalization performance of all methods in detail, we analyze the testing accuracy of the personalized model owned by each client (Figure~\ref{fig:distribution}). Both FedAMP and HeurFedAMP have more clients with higher testing accuracy on FMNIST, EMNIST and CIFAR100. We also conduct Wilcoxon signed-rank test~\cite{wilcoxon1992individual} to compare FedAMP/HeurFedAMP against the other methods on FMNIST, EMNIST and CIFAR100, a pair on a data set at a time. In all those tests, the $p$-values are all less than $10^{-4}$ and thus the non-hypotheses are all rejected.  FedAMP and HeurFedAMP outperform the other methods in testing accuracies of individual clients with statistical significance.

\begin{figure}[t]
    \centering
    \begin{subfigure}[b]{0.21\textwidth}
        \centering
        \includegraphics[width=\textwidth]{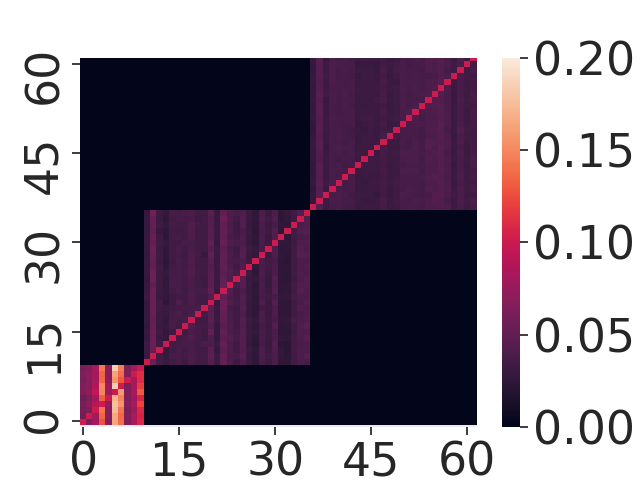} 
        \caption{FedAMP}
        \label{subfig:cluster-1}
    \end{subfigure}
    \begin{subfigure}[b]{0.21\textwidth}
        \centering
        \includegraphics[width=\textwidth]{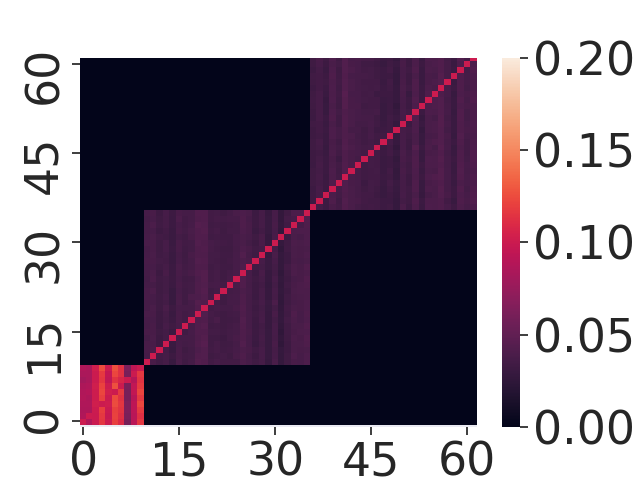}
        \caption{HeurFedAMP}
        \label{subfig:cluster-2}
    \end{subfigure}
    \caption{The visualization of the collaboration weights $\xi_{i,j}$ computed by FedAMP and HeurFedAMP on EMNIST under the practical non-IID data setting. X-axis and y-axis show the IDs of clients.}
    \label{fig:cluster}
\end{figure}

The superior performance of {FedAMP} and {HeurFedAMP} is contributed by the attentive message passing mechanism that adaptively facilitates the underlying pair-wise collaborations between clients.
Figure~\ref{fig:cluster} the visualizes the collaboration weights $\xi_{i,j}$ computed by FedAMP and HeurFedAMP. The pair-wise collaborations between clients are accurately captured by the three blocks in the matrix, where the three ground-truth collaboration groups are clients 0-9, 10-35 and 36-61.
The other methods, however, are not able to form those collaboration groups because using a single global model cannot describe the numerate pairwise collaboration relationships between clients when the data is non-IID across different clients.

\nop{
is that comparing with the algorithms (i.e., FedAvg/FedAvg-FT, FedProx/FedProx-FT, SCAFFOLD and APFL) which compute the single global model using the fixed weights during federated training, {FedAMP} and {HeurFedAMP} can discover the underlying collaboration relationships between pairs of clients and creating the personalized cloud model by using different weights for each client. This phenomenon can be observed from Fig.~\ref{fig:cluster}, which shows the pattern of the linear combination weights $\xi_{ij}$ defined in \eqref{eq:grad-step-eq} from the round achieving the BMTA for EMNIST data set. We observe that the patterns of aggregation weights of {FedAMP} and {HeurFedAMP} are consistent with the data distribution as expected. For example, the upper-case letters group (i.e., client 10-35) only collaborates within the group while has non-collaboration with the other two groups of the clients. This implies {FedAMP} and {HeurFedAMP} have found the correct inter-client collaboration and clearly explains the advantages of constructing personalized cloud models for each client over a single cloud model for all clients. 
}

\section{Conclusions}
In this paper, we tackle the challenging problem of personalized cross-silo federated learning and develop {FedAMP} and {HeurFedAMP} that introduce a novel attentive message passing mechanism to significantly facilitate the collaboration effectiveness between clients without infringing their data privacy.
We analyze how the attentive message passing mechanism iteratively enables similar clients to have stronger collaboration than clients with dissimilar models, and empirically demonstrate that this mechanism significantly improves the learning performance.

\section*{Acknowledgements}
Yutao Huang's, Jiangchuan Liu's and Jian Pei's research is supported in part by the NSERC Discovery Grant program.
All opinions, findings, conclusions and recommendations in this paper are those of the authors and do not necessarily reflect the views of the funding agencies. Most of the work of the author Zirui Zhou was done when he was affiliated with the Department of Mathematics of Hong Kong Baptist University (HKBU) and was supported in part by an HKBU Start-up Grant.

\section*{Ethics Statement}
The ever-growing regulations and laws on protecting data privacy, such as the General Data Protection Regulation\footnote{\url{https://gdpr.eu/}} of Europe, strictly restricts user data transmission between different sources.
The restrictions on data transmission have become one of the biggest challenges for many data-intensive machine learning tasks.
To tackle this challenge, we propose {FedAMP} and {HeurFedAMP} to securely and efficiently train a high performance AI model by legally using the private data held by multiple data owners without infringing the data privacy of any data owner.

\bibliography{references}



\end{document}